\documentclass[sn-mathphys,Numbered]{sn-jnl}

\usepackage{graphicx}%
\usepackage{multirow}%
\usepackage{amsmath,amssymb,amsfonts}%
\usepackage{amsthm}%
\usepackage{mathrsfs}%
\usepackage[title]{appendix}%
\usepackage{xcolor}%
\usepackage{textcomp}%
\usepackage{manyfoot}%
\usepackage{booktabs}%
\usepackage{algorithm}%
\usepackage{algorithmicx}%
\usepackage{algpseudocode}%
\usepackage{listings}%

\theoremstyle{thmstyleone}%
\newtheorem{theorem}{Theorem}
\theoremstyle{thmstyletwo}%

\theoremstyle{thmstylethree}%

\begin{document}

\title[Article Title]{Machine learning in parameter estimation of nonlinear systems}


\author*[]{\fnm{Kaushal} \sur{Kumar}}\email{kaushal.kumar@stud.uni-heidelberg.de}

\affil[]{\orgdiv{Institute for Mathematics}, \orgname{Heidelberg University}, \orgaddress{\street{Im Neuenheimer Feld}, \city{69120 Heidelberg}, \state{BW}, \country{Germany}}}


\abstract{
Accurately estimating parameters in complex nonlinear systems is crucial across scientific and engineering fields. We present a novel approach for parameter estimation using a neural network with the Huber loss function. This method taps into deep learning's abilities to uncover parameters governing intricate behaviors in nonlinear equations. We validate our approach using synthetic data and predefined functions that model system dynamics. By training the neural network with noisy time series data, it fine-tunes the Huber loss function to converge to accurate parameters. We apply our method to damped oscillators, Van der Pol oscillators, Lotka-Volterra systems, and Lorenz systems under multiplicative noise. The trained neural network accurately estimates parameters, evident from closely matching latent dynamics. Comparing true and estimated trajectories visually reinforces our method's precision and robustness. Our study underscores the Huber loss-guided neural network as a versatile tool for parameter estimation, effectively uncovering complex relationships in nonlinear systems. The method navigates noise and uncertainty adeptly, showcasing its adaptability to real-world challenges.}

\keywords{Parameter estimation, Systems identification, Nonlinear systems, Machine learning, Huber loss}



\maketitle

\section{Introduction}\label{sec1}
Accurate parameter estimation in the field of nonlinear dynamical systems holds pivotal importance across a wide spectrum of scientific and engineering domains. These estimations provide essential cornerstones for comprehending and predicting behaviors exhibited by intricate systems characterized by nonlinear dynamics. While conventional methods have yielded valuable insights into parameter estimation, the complexities of modern systems, coupled with the presence of noise and uncertainty, call for innovative methodologies capable of addressing these challenges \cite{strogatz2019nonlinear, brunton_kutz_2022, aster2018parameter}.

This paper introduces a novel approach for parameter estimation, harnessing the capabilities of neural networks in synergy with the Huber loss function—a hybrid criterion that amalgamates the characteristics of Mean Squared Error (MSE) and Mean Absolute Error (MAE). Capitalizing on neural networks' proficiency in approximating complex functions, this methodology presents a promising avenue for capturing intricate relationships inherent in nonlinear dynamical systems. Furthermore, the integration of the Huber loss function tackles the limitations of conventional loss functions by adeptly handling outliers and fluctuations, rendering the approach robust to noise intrinsic in empirical data \cite{10.1214/aoms/1177703732}.

The primary objective of this study is to present a pioneering methodology for effectively estimating parameters that govern the behavior of intricate systems described by nonlinear ordinary differential equations (ODEs). The proposed methodology is validated through the generation of synthetic data based on a predefined function $f(x,t,p)$ that encapsulates the system's intricate dynamics. The neural network, trained using noisy time series data, converges towards parameters that optimally encapsulate the underlying dynamics, thus affirming the efficacy of the approach.

Subsequently, the efficiency of the introduced methodology is demonstrated through its application to four classical systems: damped oscillators, van der Pol oscillators, Lotka-Volterra systems, and Lorenz systems. The forthcoming sections expound upon the methodology employed, the architecture of the neural network, the formulation of the Huber loss function, and the process of synthetic data generation. Our experiments yield visual comparisons of true and estimated trajectories, accompanied by discussions on the implications of our findings. Through this exploration, we underscore the potential of the proposed methodology as a versatile tool for parameter estimation in nonlinear dynamical systems, adept at navigating the intricacies arising from noise and uncertainty.

In essence, this paper contributes to the intersection of machine learning and dynamical systems research, introducing a robust framework for parameter estimation that holds significant promise across diverse scientific and engineering applications.

\section{Earlier Approaches}
Before the rise of machine learning (ML) techniques, researchers used traditional statistical and optimization methods to estimate parameters for nonlinear systems. These methods worked well but faced challenges when dealing with complex and nonlinear relationships. One common technique was the least squares method, which involved minimizing the squared differences between observed data and model predictions. This method could be adapted for nonlinear systems by iteratively adjusting parameters to reduce the differences between predictions and data \cite{doi:10.1137/0903003, 225dd72e-23e2-30d0-ad93-6fc234f74f64}.

Another approach was maximum likelihood estimation (MLE), a statistical method that aimed to find parameter values that made the observed data most likely given the model. While MLE was commonly used for linear systems, extending it to nonlinear systems was challenging due to the complexity of the likelihood function and the need for iterative optimization techniques \cite{075e2257-5240-3f6c-bda6-e2a0ba195586}.

Researchers also turned to nonlinear optimization algorithms like the Levenberg-Marquardt, Gauss-Newton, and Nelder-Mead methods. These methods repeatedly adjusted parameter values to minimize the differences between model predictions and observed data \cite{refId0,kumar2023exploring}.

System identification techniques, which aimed to create mathematical models from input-output data, were also used. These techniques included analyzing time-domain and frequency-domain data to estimate parameters governing system behavior \cite{RAISSI2019686, chen2018neural, Karniadakis2021PhysicsinformedML, kidger2022neural}.

While some of these earlier approaches had merits, they struggled with complex dynamics and high-dimensional parameter spaces. The introduction of machine learning techniques provides a promising alternative. ML methods are skilled at capturing complex nonlinear relationships and handling large datasets. Incorporating ML into parameter estimation could potentially address the limitations of previous methods, leading to more accurate and efficient parameter estimation across various fields.

\section{Methods}

Let us consider a system of ordinary differential equations for state variables $x(t)$ characterized by parameter estimation problems
\begin{equation}
\dot{x}(t) = F(x,t,p),
\end{equation}
which depends on parameter vector $p=(p_{1},p_{2},...,p_{n})$ Besides, measurement $\zeta_{ij}$ for the state factors or increasingly broad for capacities within the states are given
\begin{equation}
    \zeta_{ij}= g_{ij}(x(t_{j}),p)+\varepsilon_{ij}
\end{equation}
which are collected at measurement times $t_{j}$, $j=1,2,...,k$,
$$t_{0} \leq t_{1}<...<t_{k}\leq t_{f},$$
over a period $[t_{0},t_{f}]$, and are assumed to be affected by a measurement error $\varepsilon_{ij}$. 
The task at hand is to find the optimal values of the unknown parameters $p$ such that the model accurately reproduces the observed process. This can be achieved by minimizing an appropriate objective function, which takes into account the measurement errors $\zeta_{ij}$. If the measurement errors are independent, Gaussian, with zero mean and variances $\sigma_{ij}^{2}$, then a suitable objective function is given by the weighted $l_{2}$ norm of the measurement errors.
\begin{equation}
    l_{2}(x,p)=  \sum_{i,j} \sigma_{ij}^{-2}\varepsilon_{ij}^{2}= \sum_{ij} \sigma_{ij}^{-2}[\zeta_{ij}-g_{ij}(x(t_{j}),p)]^{2}.
    \end{equation}
To accomplish this task, we need to use optimization algorithms that find the parameter vector $p$ and trajectory $x$ that minimize the objective function.\\

Our approach to parameter estimation is to use layered feedforward neural network with a Huber Penalty Function to estimate the parameters \cite{10.1214/aoms/1177703732}. The objective function of the Huber loss-guided neural network paradigm is given as follows:

\begin{equation}
\min_{\theta} \sum_{i=1}^{N} \rho\Bigg(\frac{\zeta_{ij}-g_{ij}(x(t_{j}),p)}{\sigma_{ij}}\Bigg),
\end{equation}

where $\rho$ is the huber loss function. The Huber penalty function is defined as follows:

\begin{equation}
\rho(z) = \begin{cases}
\frac{1}{2}z^{2} & \text{if } |z| \leq \delta, \\
\delta(|z| - \frac{1}{2}\delta) & \text{otherwise},
\end{cases}
\end{equation}

where $z$ is the residual, $\delta$ is a tuning parameter that determines the threshold for the linear and quadratic penalties, and $\rho(z)$ is the penalty assigned to the residual.  In the context of optimization, the Huber loss provides a compromise between the characteristics of Mean Squared Error (MSE) and Mean Absolute Error (MAE). It combines the benefit of reduced sensitivity to outliers (like MAE) with the smoothness and differentiability of a quadratic loss (like MSE). When used in neural network training, the Huber loss helps the optimizer adjust the network's parameters to minimize the loss function, resulting in parameter estimates that are less influenced by outliers in the data.

\subsection{Parameter Estimation Using Neural Networks for Solving Ordinary Differential Equations}

Neural networks exhibit an impressive capability to excel in parameter estimation for solving ordinary differential equations (ODEs) \cite{Calin2020}. This proficiency comes to the forefront when the ODEs adhere to well-defined conditions, encompassing crucial aspects:

\begin{enumerate}
\item \textbf{Existence of Solution}: Ensuring the presence of a viable solution.
\item \textbf{Uniqueness of Solution}: Guaranteeing the uniqueness of the derived solution.
\item \textbf{Smoothness of Solution}: Ensuring the solution's smooth behavior with respect to the initial data.
\end{enumerate}

In the context of first-order ODEs,  we consider equations of the form:
\begin{equation}
    y'(t) = f(t, y(t)), \quad y(t_{0}) = y_{0}, \quad t \in [t_{0}, t_{0} + \epsilon),
\end{equation}
where $\epsilon > 0$. It is established that when the function $f(t, \cdot)$ satisfies Lipschitz continuity in the second variable, the above ODE has a unique solution for sufficiently small $\epsilon$. For simplicity, we consider a one-dimensional function $f: [t_{0}, t_{0} + \epsilon) \times \mathbb{R} \longrightarrow \mathbb{R}$. Nevertheless, this framework extends harmoniously to multi-dimensional cases where $y \in \mathbb{R}^{m}$, leading to systems of ODEs.

Our trajectory embarks on the construction of a feedforward neural network, tailored to excel in parameter estimation. This network's input is the continuous variable $t \in [t_{0}, t_{0} + \epsilon)$, and its output is a smooth function $\phi_{\theta}(t)$ that adeptly approximates the solution $y(t)$ of the aforementioned ODE. The parameters governing the neural network are collectively represented as $\theta$.

Our journey involves formulating a cost function using the Huber loss:
\begin{equation}
    C(\theta) = \sum_{k=0}^{n-1} \text{Huber}\left(\frac{\phi_{\theta}(t_{k+1}) - \phi_{\theta}(t_{k})}{\Delta t} - f(t_{k}, \phi_{\theta}(t_{k})), \delta\right) \Delta t + \text{Huber}(\phi_{\theta}(t_{0}) - y_{0}, \delta),
\end{equation}

where $\text{Huber}(x, \delta)$ embodies the Huber loss function, with $\delta$ serving as a tuning parameter. This loss function fortifies the model's resilience against outliers present in the data.

The architectural setup comprises a neural network endowed with a single hidden layer, housing $N$ hidden neurons. The output layer employs a linear activation function. The ensuing expression captures the essence of this configuration:

\begin{equation}
\phi_{\theta}(t) = \sum_{j=1}^{N} \alpha_{j} \sigma (w_{j}t + b_{j}), \quad \theta = (w, b, \alpha) \in \mathbb{R}^{N} \times \mathbb{R}^{N} \times \mathbb{R}^{N}.
\end{equation}

The differentiability of $\phi_{\theta}(t)$ is unequivocally affirmed due to the inherent differentiability of the logistic function $\sigma$. At this crossroads, two pathways beckon:

\begin{enumerate}
\item \textbf{Incorporation of Relation}: We weave Equation (8) into the empirical cost function $C(\theta)$, enriching it with the Huber loss. Minimization of this function becomes a reality through the implementation of the Adam optimizer. This path leads us to $\phi_{\theta^{*}}(t)$, a close approximation of the sought-after solution $y(t)$.
\item \textbf{Gradient Computation}: The computation of gradients is performed in tandem with the Huber loss and the relation:
\end{enumerate}

\begin{align*}
\nabla_{\theta}C(\theta) &= \sum_{k=0}^{n-1} \text{HuberGradient}\left(\frac{\phi_{\theta}(t_{k+1}) - \phi_{\theta}(t_{k})}{\Delta t} - f(t_{k}, \phi_{\theta}(t_{k})), \delta\right) \nabla_{\theta}\phi_{\theta}(t_{k}) \Delta t \\
&\quad + \text{HuberGradient}(\phi_{\theta}(t_{0}) - y_{0}, \delta) \nabla_{\theta}\phi_{\theta}(t_{0}),
\end{align*}

where $\text{HuberGradient}(x, \delta)$ designates the derivative of the Huber loss function with respect to $x$. By employing the Adam optimizer, we forge an iterative approximation sequence:
\begin{equation}
    \theta_{j+1} = \text{AdamUpdate}(\theta_{j}, \nabla_{\theta}C(\theta)),
\end{equation}

where $\text{AdamUpdate}$ encapsulates the update rule intrinsic to the Adam optimizer. Through these strategic paths, neural networks emerge as stalwart tools for robust parameter estimation, unveiling the intricate tapestry of solutions within ordinary differential equations.

\subsection{Multi-Layer Perceptrons (MLPs)}

\begin{figure}[h]%
\centering
\includegraphics[width=0.7\textwidth]{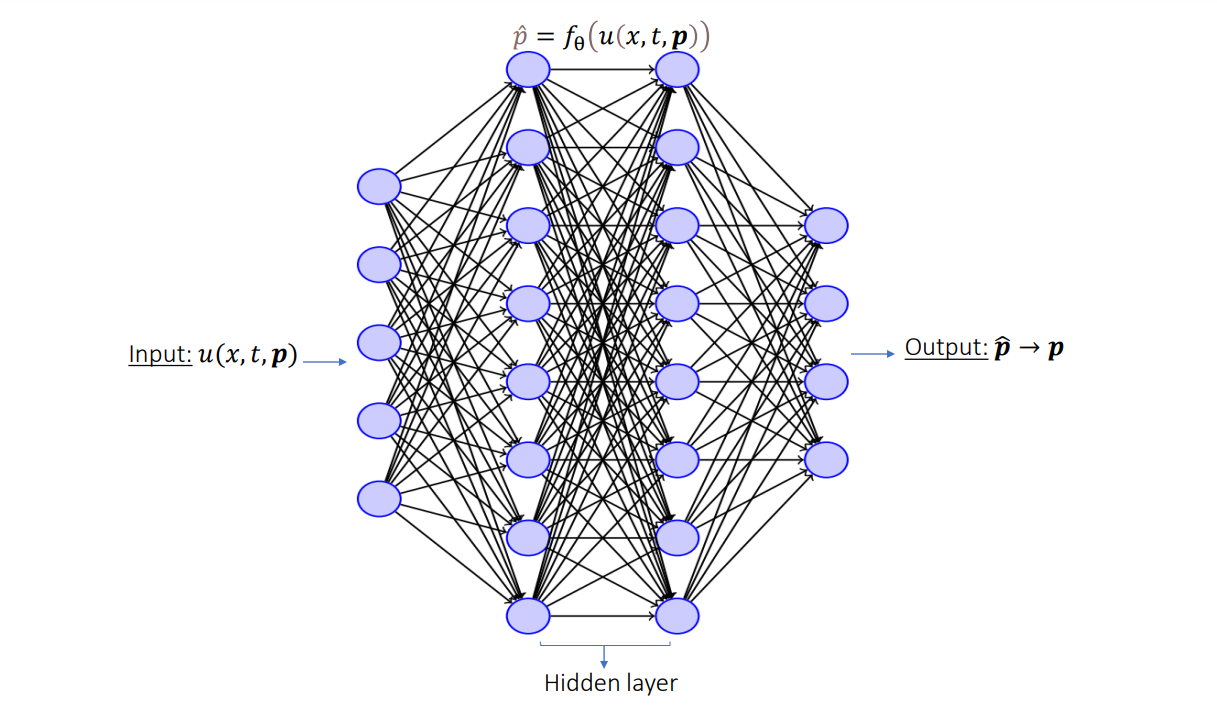}
\caption{MLP architecture}\label{mlp}
\end{figure}

Multi-Layer Perceptrons (MLPs) are supervised learning algorithms designed to understand complex relationships \(f(\cdot): \mathbb{R}^{n} \longrightarrow \mathbb{R}^{m}\). They learn from datasets where \(n\) is the input space dimension and \(m\) is the output space dimension. By using features \(X = x_{1}, x_{2}, \ldots, x_{n}\) and corresponding target values \(y\), MLPs create non-linear approximations suitable for tasks like classification and regression \cite{Haykin2010NeuralNA}. MLPs consist of input and output layers, as well as hidden layers that capture intricate data patterns and relationships, as shown in Figure \ref{mlp}.

An MLP, also known as a feedforward neural network, comprises layers of interconnected neurons. Neurons within each layer connect to all neurons in the previous and next layers. This architecture empowers MLPs to effectively grasp complex non-linear functions and data relationships.

Mathematically, each network layer's computations can be summarized as follows:

Given a time series training dataset \((x_{1}, y_{1}), (x_{2}, y_{2}), \ldots, (x_{n}, y_{n})\) where \(x_{i} \in \mathbb{R}^{n}\), a single hidden neuron in an MLP layer learns \(f(x) = W_{2}g(W_{1}^{T}x + b_{1}) + b_{2}\). Here, \(W_{1} \in \mathbb{R}^{m}\) and \(W_{2}, b_{1}, b_{2} \in \mathbb{R}\) are model parameters. \(W_{1}, W_{2}\) are weights, and \(b_{1}, b_{2}\) are biases. \(g(\cdot): \mathbb{R}\longrightarrow \mathbb{R}\) is the activation function.

\textbf{Hidden Layers:}
Neurons in hidden layers compute weighted inputs, apply an activation function, and pass results to the next layer. The calculation for a hidden neuron is:
\begin{equation}
    z_j = \sum_{i} w_{ij}x_i + b_j
\end{equation}
\begin{equation}
    a_j = \text{ReLU}(z_j)
\end{equation}
Where \(z_j\) sums weighted inputs, \(w_{ij}\) are weights, \(b_j\) is a bias, and ReLU is the activation function.

\textbf{Output Layer:}
The output layer predicts. It works like hidden layers but is tailored for the task. Here, it predicts parameters $[p_{1}, p_{2},..., p_{k}]$. The network's output is $\mathbf{y}_{\text{pred}} =[p_{1_{\text{pred}}}, p_{2_{\text{pred}}},..., p_{k_{\text{pred}}}]$

\textbf{Training and Optimization:}
Training adjusts weights and biases using optimization (e.g., Adam). The network compares predictions $\mathbf{y}_{\text{pred}}$ to actual $[p_{1}, p_{2},..., p_{k}]$ using a loss function, like Huber loss. The optimizer refines weights and biases iteratively to minimize the loss.

\textbf{Learning Process:}
Through iterative steps, the network learns to emulate the underlying function connecting input time series data and function parameters $f(t, y, p)$.

\begin{algorithm}
\caption{Forward Propagation in a Multi-Layer Perceptron (MLP)}\label{algo1}
\begin{algorithmic}[1]
\Require Input pattern $\mathbf{x}$, MLP architecture, list of neurons in topological order
\Ensure Calculate the MLP's output
\For{each input neuron $i$}
\State Assign $a_i$ as $x_i$
\EndFor
\For{each hidden and output neuron $i$ in topological order}
\State Compute $net_i$ as $w_{i0} + \sum_{j \in Pred(i)} w_{ij} a_j$
\State Evaluate $a_i$ using the activation function: $a_i = f_{\text{activation}}(net_i)$
\EndFor
\For{each output neuron $i$}
\State Construct the output vector $\mathbf{y}$ by gathering $a_i$
\EndFor
\State \textbf{Return} $\mathbf{y}$
\end{algorithmic}
\end{algorithm}

In summary, neural networks exploit linear input integration, non-linear activation functions, and advanced optimization techniques to approximate complex input-output relationships (see Algorithm \ref{algo1}). This adaptability positions them as versatile tools for diverse tasks including parameter estimation, regression, classification, and beyond.

\subsection*{Activation Functions in Neural Networks}
\begin{figure}[h]%
\centering
\includegraphics[width=0.7\textwidth]{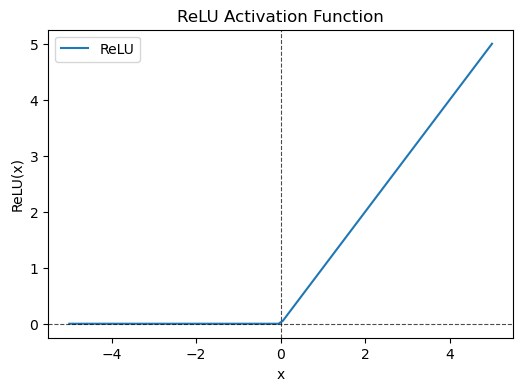}
\caption{Rectified Linear Unit (ReLU) Activation Function}\label{relu}
\end{figure}

Activation functions are vital in neural networks, determining how individual neurons respond based on their inputs. Choosing the right activation function is crucial for solving specific problems. In our study, we explored various activation functions to find the most suitable ones. Among them, the Rectified Linear Unit (ReLU) stands out. ReLU is widely used for introducing non-linearity into neural networks. It directly outputs positive inputs and converts non-positive values to zero. Mathematically, ReLU is defined as:
\begin{equation}
f(x)=\max(0,x)
\end{equation}
For a visual understanding, see Figure \ref{relu}, illustrating the distinctive non-linear behavior of the ReLU function.

\section*{Adam Optimizer}
The optimization of neural network models is a fundamental undertaking in machine learning. One of the prominent optimization algorithms is the Adam optimizer \cite{kingma2017adam}, which stands for Adaptive Moment Estimation. Adam combines the merits of both the Adagrad and RMSProp optimizers by adaptively adjusting learning rates for each parameter. This adaptability is achieved through the computation of exponential moving averages of both the gradients' first moments (mean) and second moments (uncentered variance). Mathematically, the Adam optimizer updates the parameters $\theta$ of the model with the following equations:\\
\textbf{Compute Moments:}
\begin{eqnarray}
    m_t &= \beta_1 \cdot m_{t-1} + (1 - \beta_1) \cdot g_t \nonumber \\
v_t &= \beta_2 \cdot v_{t-1} + (1 - \beta_2) \cdot g_t^2
\end{eqnarray}
where $g_t$ is the gradient of the objective function with respect to $\theta$ at time step $t$, and $m_t$ and $v_t$ are the first and second moments estimates.\\
\textbf{Bias Correction:}
\begin{eqnarray}
    \hat{m}_t &= \frac{m_t}{1 - \beta_1^t} \nonumber \\
\hat{v}_t &= \frac{v_t}{1 - \beta_2^t}
\end{eqnarray}
These bias-corrected moments counteract the initialization bias during the initial steps.\\
\textbf{Update Parameters:}
\begin{equation}
    \theta_{t+1} = \theta_t - \frac{\eta}{\sqrt{\hat{v}_t} + \epsilon} \cdot \hat{m}_t
\end{equation}
Here, $\theta_{t+1}$ represents the updated parameters, $\eta$ is the learning rate, and $\epsilon$ is a small constant added for numerical stability.

The hyperparameters $\beta_1$ and $\beta_2$ control the decay rates of the moment estimates, and $\epsilon$ prevents division by zero. Typically, default values of $\beta_1 = 0.9$, $\beta_2 = 0.999$, and $\epsilon = 10^{-8}$ are used. The Adam optimizer's adaptability, efficiency, and reliable performance across various tasks have positioned it as a widely employed optimization algorithm in training neural networks.

\subsection*{Function Approximation in Neural Networks}

Neural networks possess a remarkable property: they can accurately approximate a wide range of functions within specific function spaces. This is achieved through iterative learning, where the network gradually adjusts its outputs to match a desired target function \cite{Haykin2010NeuralNA, Goodfellow-et-al-2016}.

This approximation capability is highlighted by neural networks' ability to approximate any function in a compact set, as long as the activation function is not a polynomial. This property can be understood through the density of the set
\begin{equation}
\Sigma_{d} (\sigma) = \text{span} { \sigma(w \cdot x + b) : w \in \mathbb{R}^{d}, b \in \mathbb{R} }
\end{equation}
in the space $C(\Omega)$, where $\Omega$ is a compact subset of $\mathbb{R}^{d}$.\\

The Universal Approximation Theorem is fundamental in the realm of neural networks and function approximation. It states that a neural network with a single hidden layer and a sufficient number of hidden units can approximate any continuous function over a compact domain with arbitrary precision. This theorem showcases the power and versatility of neural networks.\\

\begin{theorem}
\textbf{Universal Approximation Theorem}:-\
For a continuous function $f: [a, b] \longrightarrow \mathbb{R}$ and an error tolerance $\epsilon > 0$, there exists a single hidden layer feedforward neural network with a finite number of neurons that approximates $f$ to the extent that:
\begin{equation}
    \sup_{x\in [a,b]} \|f(x) -\hat{f}(x)\| <\epsilon
\end{equation}
where $\hat{f}$ is the network's output.
\end{theorem}

The proof of the Universal Approximation Theorem involves principles from measure theory, functional analysis, and approximation theory. George Cybenko first introduced this theorem in 1989 \cite{Cybenko1989ApproximationBS}, focusing on sigmoidal activation functions. Subsequent work expanded its scope to various activation functions and network architectures.

For a thorough proof of the Universal Approximation Theorem, one can refer to resources like Cybenko's original paper ("Approximation by Superpositions of a Sigmoidal Function"), along with other texts on neural networks and approximation theory \cite{Hornik1991ApproximationCO, lu2017expressive}.

\section{Results and Analysis}\label{sec2}
In this section, we present the results and analysis of our study, which focuses on evaluating the robustness and accuracy of the methods discussed in Section 3. Our simulations encompassed various systems of different complexities, subject to multiplicative noise with different intensities \cite{Gottwald2013TheRO, oksendal1998, haunggi1994colored,doi:10.1137/0148023}. We delved into the effects of both Gaussian and colored noise separately to provide a comprehensive understanding of their impact.

To examine noise's influence, we introduced uncorrelated noise 
$\eta(t)$ characterized by specific properties:
\begin{equation}
\langle\eta(t)\rangle = 0 \quad \text{and} \quad \langle\eta(t)\eta(t')\rangle = \delta(t - t').
\end{equation}

where \(\langle\cdot\rangle \) indicates temporal averaging. We then incorporated this noise into the equation of motion.

Our investigation involved two types of noise: white Gaussian noise and colored noise. White Gaussian noise, uniform across frequencies, emulates random measurement errors and uncertainties present in real data. The addition of this noise can affect parameter estimation accuracy, which tends to degrade with increasing noise intensity. On the other hand, colored noise exhibits varying power levels across frequencies, posing challenges for parameter estimation due to its complex frequency distribution. Our study encompassed both white noise and pink noise ($1/f$).

For each system under consideration, we generated noisy data through simulation using known parameters. Gaussian noise was then added to the output. Subsequently, optimization algorithms were employed to estimate the system's parameters from the noisy data. This process was repeated 10 times for both Gaussian and colored (pink) noise scenarios. We computed average values for the estimated parameters and Huber loss, a metric quantifying solution accuracy.

Our analysis using Huber-guided neural networks demonstrated the successful estimation of parameters across various systems under different conditions. This approach proved effective even when the systems were subjected to different levels of noise intensities, represented as $\eta= [0.0001, 0.001, 0.01, 0.1]$. Notably, these neural networks consistently captured the intricate behaviors of the systems, including damped harmonic oscillators, van der Pol oscillators, Lotka-Volterra models, and the Lorenz system. Even in the presence of varying levels of Gaussian noise, these networks maintained their robustness. The estimated parameter values closely aligned with the actual values, and the optimized paths faithfully portrayed the systems' behaviors.

Importantly, our results were in line with the Sparse Identification of Nonlinear Dynamical Systems (SINDY) techniques proposed by Brunton et al. (2016) \cite{doi:10.1073/pnas.1517384113} and optimization methods presented by Kumar et al. (2023) \cite{refId0}.

The crux of our analysis lay in comparing the effects of Gaussian and colored (pink) noise on parameter estimation and solution accuracy. Through various simulations and tabulated results, we revealed subtle variations in estimated parameter values between the two noise types. This implied that noise characteristics could introduce bias, underlining the necessity of accounting for noise attributes. Our exploration of Huber loss consistently demonstrated comparable accuracy between Gaussian and colored noise, with the latter exhibiting a slight advantage. This observation underscores the significance of noise characteristics in refining estimation outcomes.

\subsection{Test Problem: Two-dimensional Damped Oscillator}
The dynamics of the two-dimensional damped harmonic oscillator are described by the following equations:
\begin{eqnarray}
\frac{dx_{1}}{dt}=p_{1}x_{1}^{3}+p_{2}x_{2}^{3}\nonumber\\ \label{eq7}
\frac{dx_{2}}{dt}=p_{3}x_{1}^{3}+p_{4}x_{2}^{3}
\end{eqnarray}

Here, $x_{1}$ and $x_{2}$ represent the state variables, while $p_{1}, p_{2}, p_{3}$, and $p_{4}$ denote the system's parameters. For our study, the true parameter values are $p_{1}=-0.1, p_{2}=2, p_{3}=-2$, and $p_{4}=-0.1$. The initial conditions are set at $[x_{1_{0}},x_{2_{0}}]^{T}=[2.0,0.0]^{T}$.

\begin{table}[ht]
\caption{Parameter estimations across various noise levels (damped oscillator)}\label{table1} 
\label{tab:estimated_parameters}
\centering
\begin{tabular}{ccccc}
\hline
Noise Level & $\hat{p_{1}}$ & $\hat{p_{2}}$ & $\hat{p_{3}}$ & $\hat{p_{4}}$  \\
\hline
0.0001 & -0.0996 & 2.0036 & -2.0029 & -0.0998  \\
0.001 & -0.0999 & 2.0001 & -2.0002 & -0.0999\\
0.01 & -0.0979 & 1.9961 & -1.9941 & -0.0971 \\
0.1 & -0.1002 & 1.9988 & -1.9960 & -0.1003 \\
\hline
\end{tabular}
\end{table}
Table \ref{table1} provides the estimated parameter values under various levels of Gaussian noise. It is evident that even amidst noise, the estimated parameters closely align with the true parameters, demonstrating the effectiveness of our methodology.

\begin{figure}[h]%
\centering
\includegraphics[width=0.70\textwidth]{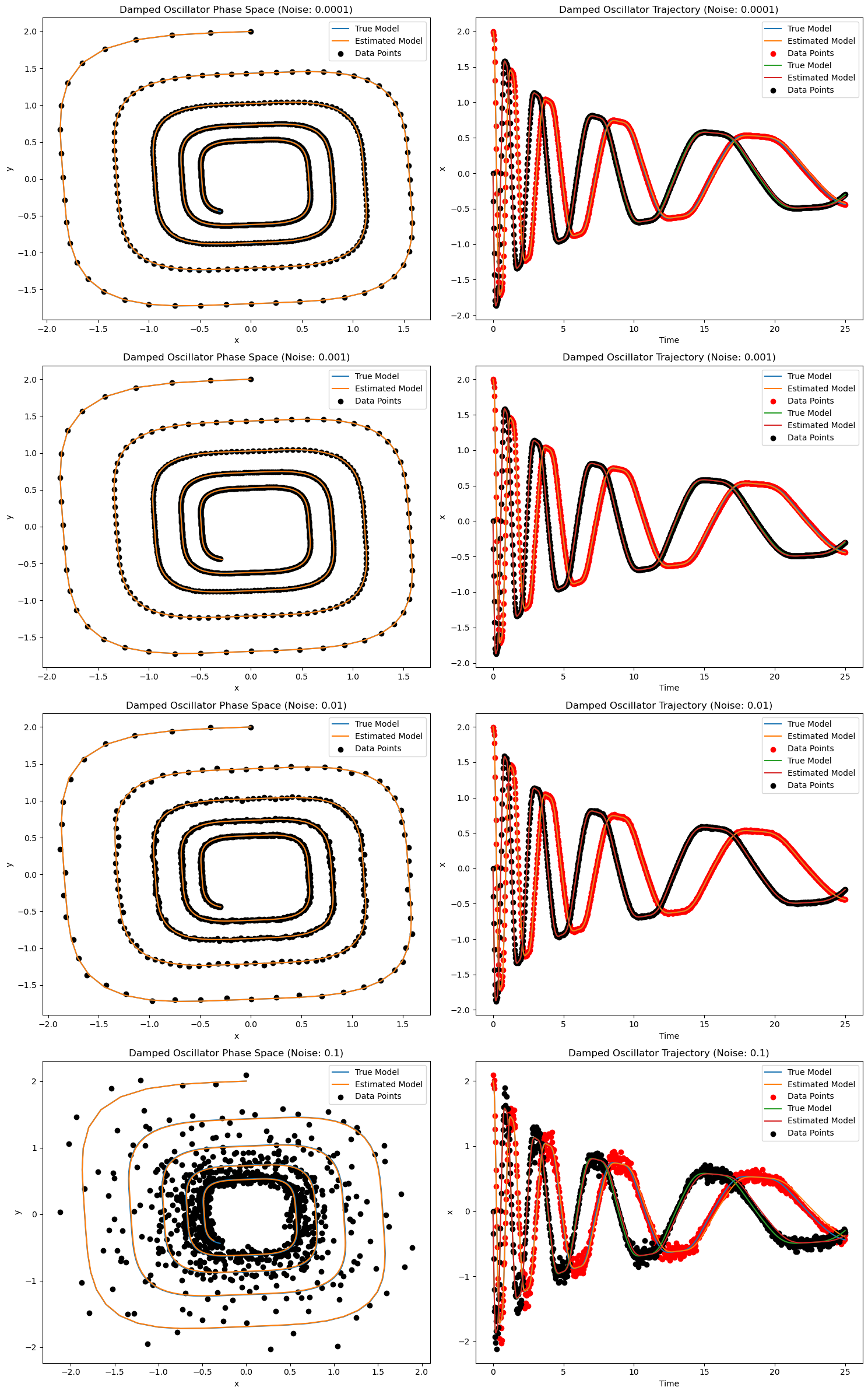}
\caption{The identified system adeptly captures the dynamics of the two-dimensional damped harmonic oscillator featuring cubic dynamics. In the visual representation, solid colored lines correspond to the true system dynamics, whereas dashed lines portray the learned dynamics. The phase portrait showcases the accurate reproduction of the system's behavior.}\label{fig1}
\end{figure}

\begin{table}[h!]
\caption{Impact of Noise Characteristics on Parameter Estimation and Solution Accuracy: A Comparative Analysis between Gaussian and Colored (Pink) Noise}\label{table2}
\centering
\begin{tabular}{ccc}
\hline
  & Gaussian Noise & Colored (Pink) Noise \\
\hline
$\hat{p_{1}}$ & -0.1001 & -0.1005 \\
$\hat{p_{2}}$ & 1.9998 & 2.0003 \\
$\hat{p_{3}}$ & -1.9999 & -2.0007 \\
$\hat{p_{4}}$ & -0.0998 & -0.1004 \\
Huber Loss & 0.000000133 & 0.000000269\\
\hline
\end{tabular}
\end{table}
Table \ref{table2} presents a comparative analysis between Gaussian noise and colored (pink) noise, showcasing the impact of noise characteristics on parameter estimation and solution accuracy. Remarkably, the estimated parameters show consistent alignment with true values across both noise types. Figure \ref{fig1} visually represents the accuracy of Huber-guided neural networks in capturing the dynamics and phase portraits of the two-dimensional damped harmonic oscillator under varying noise levels. The solid colored lines depict the true system dynamics, while the dashed lines illustrate the learned dynamics. The phase portrait vividly illustrates the precision with which our method reproduces the system's behavior. 

Through this analysis of the two-dimensional damped harmonic oscillator, we confirm the robustness and efficacy of our approach in parameter estimation and solution accuracy, even in the presence of noise.

\subsection{Test Problem: van der Pol Oscillator}
The equation governing the van der Pol oscillator is defined as:
\begin{equation}
\frac{d^2x}{dt^2}-\mu(1-x^2)\frac{dx}{dt}+x=0
\end{equation}

Here, $x$ denotes the oscillator's displacement, $t$ represents time, and $\mu$ controls the nonlinearity \cite{1084738}. Alternatively, this equation can be represented as a pair of coupled first-order equations:

\begin{eqnarray}
\frac{dx_1}{dt} &=& x_2 \nonumber \\
\frac{dx_2}{dt} &=& \mu(1 - x_1^2)x_2 - x_1.
\end{eqnarray}
For our analysis, we initialize the system with $[x_{1_{0}} \hspace{0.1cm} x_{2_{0}}]^{T} = [2.0 \hspace{0.2cm} 0.0]^{T}$ and utilize a time step size of $\delta t = 0.01$

\begin{figure}[h]%
\centering
\includegraphics[width=0.85\textwidth]{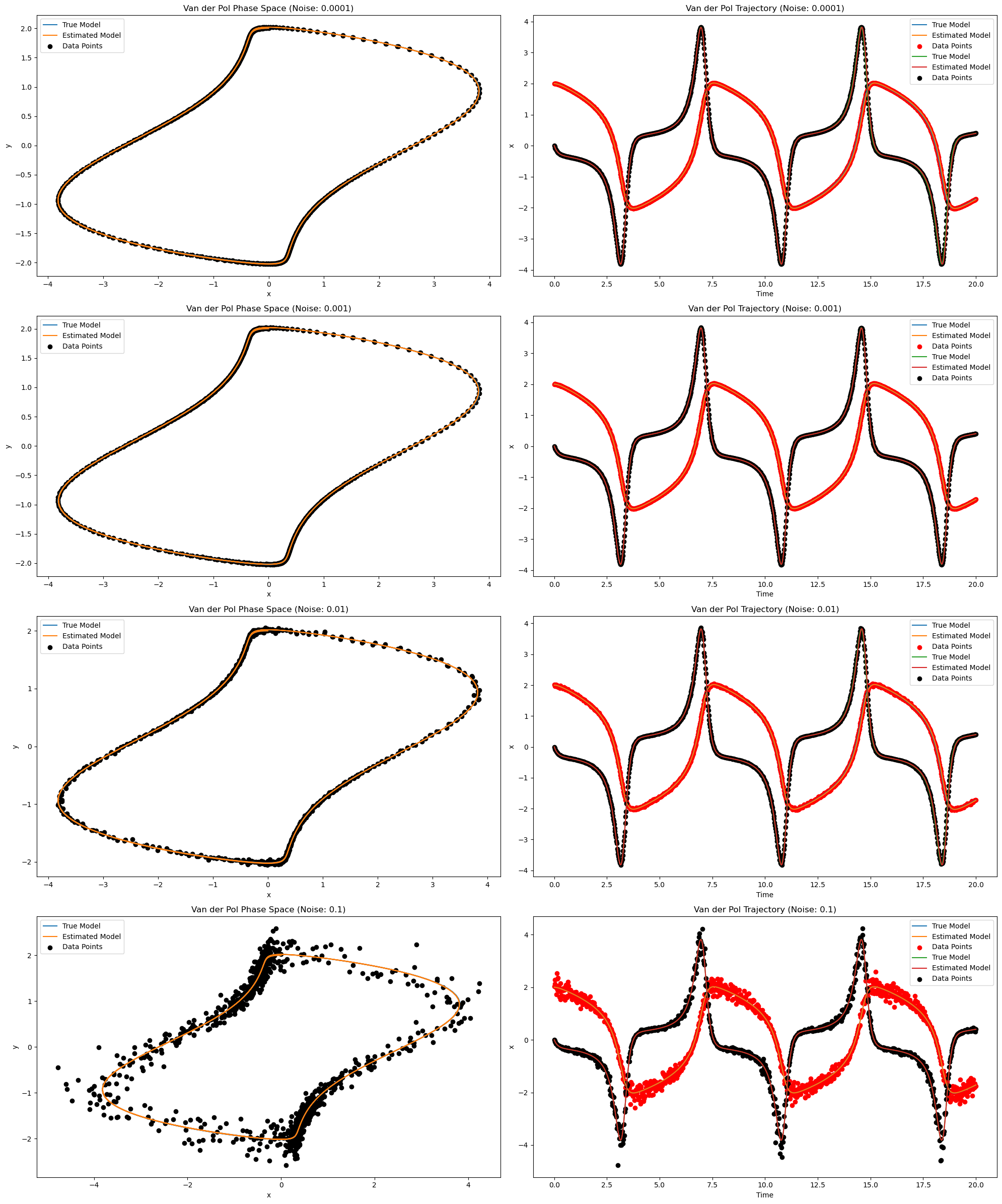}
\caption{In the context of the Van der Pol oscillator, the Huber-guided neural networks method successfully replicates both trajectories on the left side and phase portraits on the right side. The chosen initial condition is given by $[x_{1_{0}} \ x_{2_{0}}]^{T} = [2.0 \ 0.0]^{T}$. Our comparison involves evaluating the resultant trajectories against both the ground truth trajectories and the data perturbed by noise.}\label{fig2}
\end{figure}

\begin{table}[h!]
\caption{Parameter estimations across various noise levels (van der Pol Oscillator)}\label{table3}
\centering
\begin{tabular}{ccc}
\hline
Noise Level & True Parameter & Estimated Parameter \\
\hline
0.0001 & 2.0 & 2.0006\\
0.001 & 2.0 & 2.0012 \\
0.01 & 2.0 & 1.9967 \\
0.1 & 2.0 & 1.9938 \\
\hline
\end{tabular}
\end{table}
Table \ref{table3} showcases the results of parameter estimations across various noise levels for the van der Pol oscillator. Notably, the Huber-guided neural networks accurately estimate parameters that closely align with the true values.
Figure \ref{fig2} further illustrates the accuracy of our method by comparing the estimated trajectories and phase portraits with both true trajectories and noisy data. This visual comparison demonstrates the precision of the Huber-guided neural networks method.
\begin{table}[h!]
\caption{Impact of Noise Characteristics on Parameter Estimation and Solution Accuracy: A Comparative Analysis between Gaussian and Colored (Pink) Noise}\label{table4}
\centering
\begin{tabular}{ccc}
\hline
  & Gaussian Noise & Colored (Pink) Noise \\
\hline
$\hat{\mu}$ & 1,9999 & 2,0001 \\
Huber Loss & 0.0000208 & 0.0000183 \\
\hline
\end{tabular}
\end{table}
Table \ref{table4} offers a comparative analysis between Gaussian noise and colored (pink) noise. The results underscore the method's consistent performance across both noise types. Huber loss values provide a quantification of accuracy, confirming the advantageous influence of colored noise.

The application of Huber-guided neural networks to the van der Pol oscillator showcases the method's efficacy in reproducing trajectories and phase portraits, even in the presence of noise.

\subsection{Test problem: Lotka-Volterra Model}
The Lotka-Volterra equations are defined as:
\begin{align}
\frac{dx}{dt} &= \alpha x - \beta xy \nonumber \\
\frac{dy}{dt} &= \delta xy - \gamma y
\end{align}
Here, $x$ represents the prey population, $y$ signifies the predator population, and \(\alpha\), \(\beta\), \(\gamma\), and \(\delta\) are parameters that influence system dynamics \cite{wangersky1978lotka}. These equations capture cyclic interactions inherent in ecosystems, often leading to oscillatory patterns.

For our analysis, we set the true parameter values as \(\alpha = 1.0\), \(\beta = 0.5\), \(\gamma = 0.5\), and \(\delta = 2.0\), with initial conditions $[x_{0},y_{0}]=[2.0,1.0]$. Employing the Huber loss as a guiding optimization criterion, we trained a neural network model to achieve reliable parameter estimates. The accuracy of these estimates remained robust across varying noise intensities, as demonstrated in Table \ref{table5}.

\begin{figure}[h]
    \centering
    \includegraphics[width=0.71\textwidth]{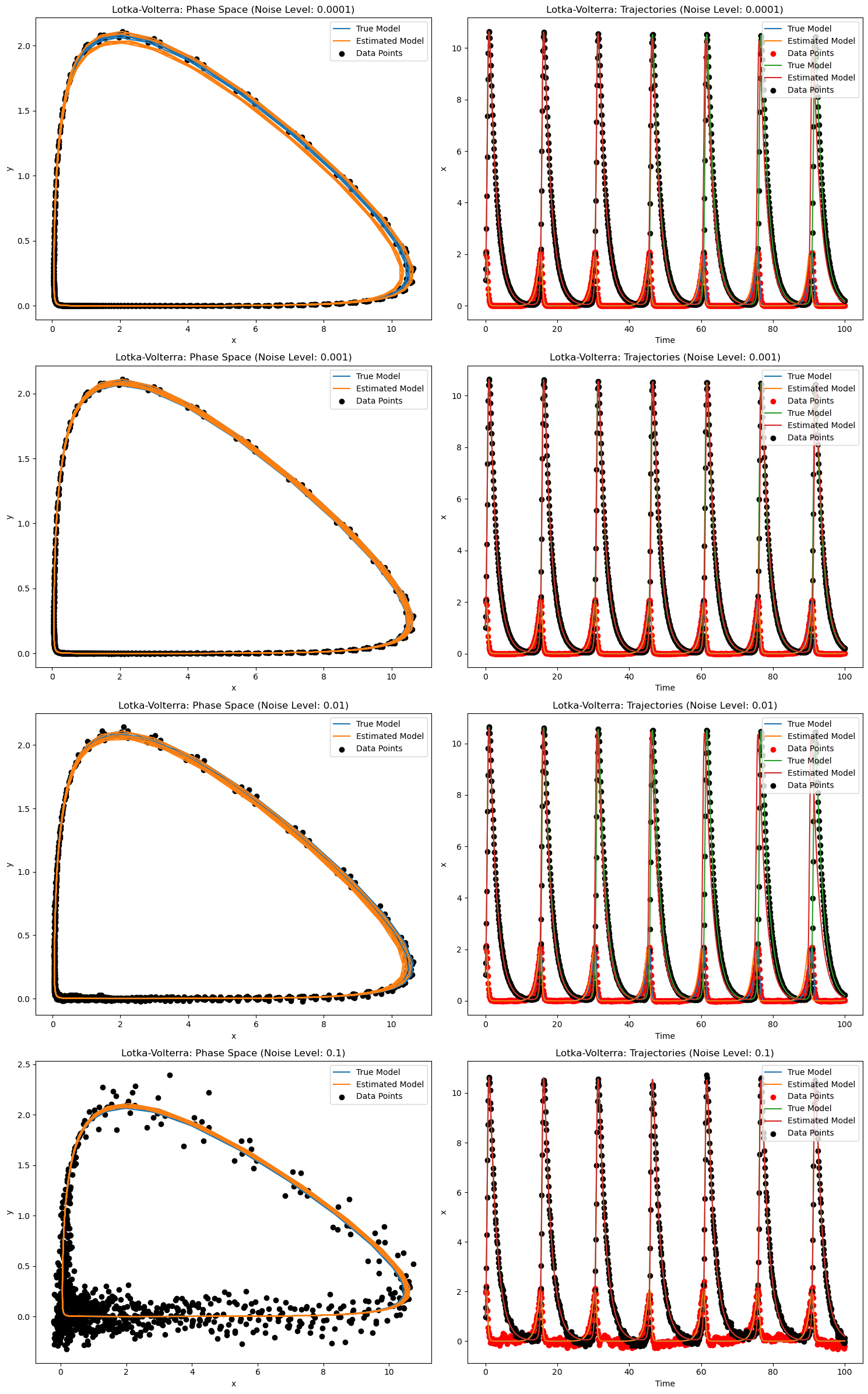}
    \caption{ Comparison of True and Estimated Trajectories of the Lotka-Volterra Model. The figure showcases the dynamic trajectories of the Lotka-Volterra model, depicting the true trajectories (blue solid lines) and the trajectories estimated through the Huber-guided neural network ( red lines).}
    \label{fig3}
\end{figure}

\begin{table}[ht]
\caption{Parameter estimations across various noise levels (Lotka-Volterra Model)}\label{table5}
\centering
\begin{tabular}{ccccc}
\hline
Noise Level & \(\hat{\alpha}\) & \(\hat{\beta}\) & \(\hat{\gamma}\) & \(\hat{\delta}\)  \\
\hline
0.0001 & 1.0008 & 0.5010 & 0.4973 & 1.9990  \\
0.001 & 1.0006 & 0.5006 & 0.5012 & 2.0005\\
0.01 & 0.9987 & 0.5001 & 0.4985 & 2.0002 \\
0.1 & 0.9995 & 0.4999 & 0.4993 & 1.9998 \\
\hline
\end{tabular}
\end{table}
Figure \ref{fig3} presents a visual comparison between true and estimated trajectories of the Lotka-Volterra model. The close alignment between the two sets of trajectories highlights the accuracy of our proposed neural network-based parameter estimation method for capturing the behavior of nonlinear systems like the Lotka-Volterra model.

\begin{table}[h!]
\caption{Impact of Noise Characteristics on Parameter Estimation and Solution Accuracy: A Comparative Analysis between Gaussian and Colored (Pink) Noise}\label{table6}
\centering
\begin{tabular}{ccc}
\hline
  & Gaussian Noise & Colored (Pink) Noise \\
\hline
\(\hat{\alpha}\) & 0.9986 & 0.9998 \\
\(\hat{\beta}\) & 0.4998 & 0.4997 \\
\(\hat{\gamma}\) & 0.4999 & 0.5000 \\
\(\hat{\delta}\) & 1.9982 & 1.9997 \\
Huber loss & 0.000119 & 0.000102\\
\hline
\end{tabular}
\end{table}

Table \ref{table6} offers a comparative analysis between Gaussian noise and colored (pink) noise, further illustrating the impact of noise characteristics on parameter estimation and solution accuracy. The estimated parameters remain consistent across both noise types, with improved accuracy observed for colored noise. The Lotka-Volterra model serves as a complex yet illuminating test case for our parameter estimation approach. Our results underscore the method's effectiveness in accurately estimating parameters and capturing system dynamics under varying noise scenarios. This reaffirms the method's potential for applications in ecological and biological studies.

\subsection{Test problem: Lorenz System}
The Lorenz system, a set of three nonlinear ordinary differential equations that gained prominence through Edward Lorenz's work in the 1960s \cite{DeterministicNonperiodicFlow}. This system offers insights into the behavior of three variables, namely $x_{1}$, $x_{2}$, and $x_{3}$, over time. The Lorenz system's equations are given by:
\begin{align}
\frac{dx_{1}}{dt} &= \sigma(x_{2} - x_{1})\nonumber\\
\frac{dx_{2}}{dt} &= x_{1}(\rho - x_{3}) - x_{2}\nonumber\\
\frac{dx_{3}}{dt} &= x_{1}x_{2} - \beta x_{3}
\end{align}
Where $\sigma$, $\rho$, and $\beta$ are the governing parameters. We conducted simulations by numerically integrating these equations with initial conditions $[x_{1_{0}}, \ x_{2_{0}}, \ x_{3_{0}}]^{T} = [-8, \ 7, \ 27]^{T}$, over a time span from $t = 0$ to $t = 25$, utilizing a time step size of $\Delta t = 0.01$, and add varying levels of noise. The true parameter values were $\sigma = 10.0$, $\rho = 28.0$, and $\beta = 8/3$.
\begin{table}[ht]
\caption{Parameter estimations across various noise levels (Lorenz system)}\label{table7}
\label{tab:estimated_parameters}
\centering
\begin{tabular}{cccc}
\hline
Noise Level & $\hat{\sigma}$ & $\hat{\rho}$ & $\hat{\beta}$ \\
\hline
0.0001 & 9.9975 & 27.9929 & 2.6654 \\
0.001 & 9.9985 & 27.9985 & 2.6661 \\
0.01 & 9.9999 & 28.0002 & 2.6663 \\
0.1 & 10.0001 & 27.9991 & 2.6668 \\
\hline
\end{tabular}
\end{table}
Utilizing Huber-guided neural networks, we estimated the Lorenz system's parameters from noisy data. Table \ref{table7} displays the estimated parameter values under different noise levels. Notably, even in the presence of noise, the estimated parameters closely align with the true values.
\begin{figure}
\resizebox{1.0\columnwidth}{!}{
  \includegraphics{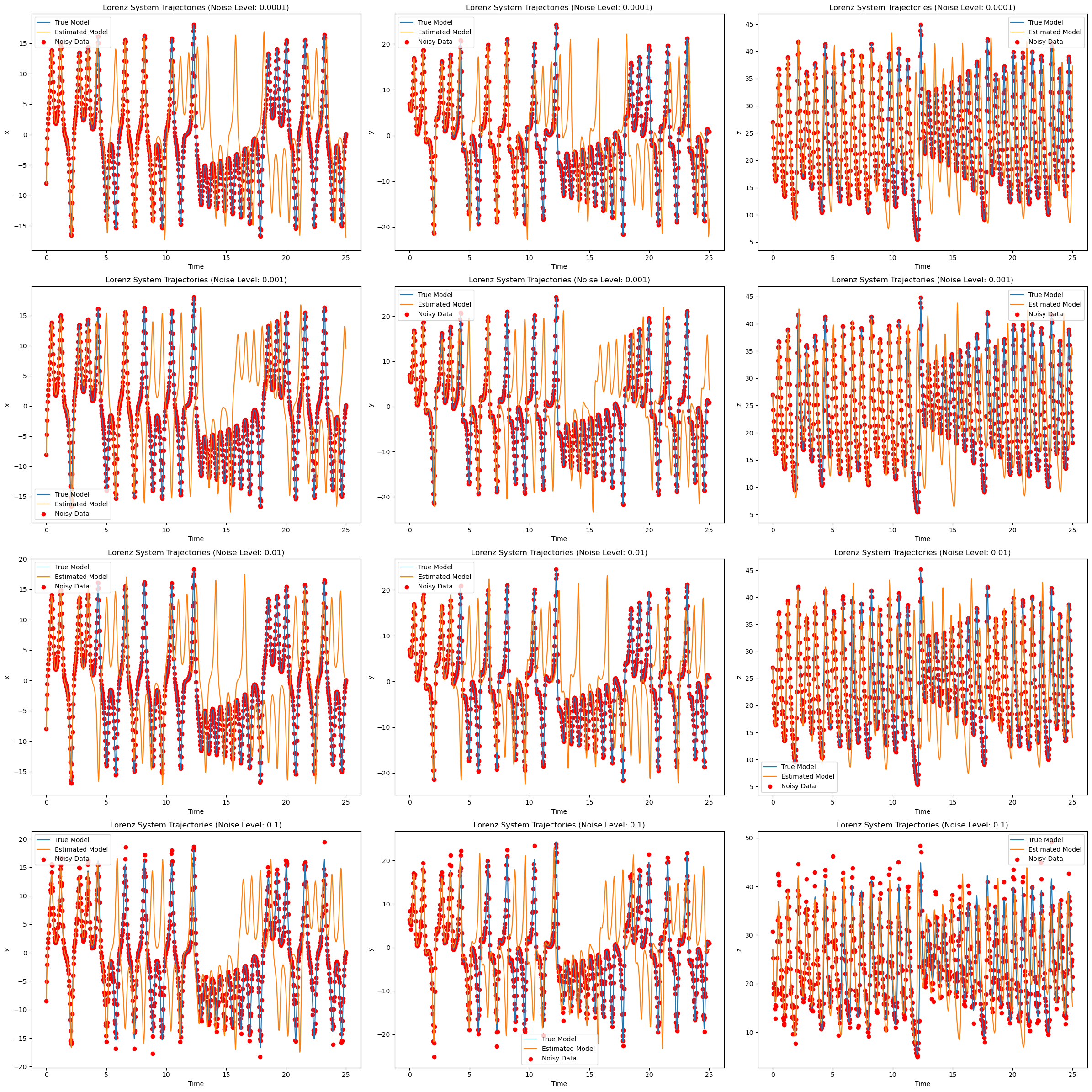} }
\caption{We analyze the dynamic trajectories of the Lorenz system, with a specific emphasis on scenarios where measurements of position ($x$) and velocity ($\Dot{x}$) are influenced by noise. The true system trajectories are represented by solid blue lines, while the estimated trajectories, obtained through the application of neural networks, are depicted using dashed red arrows.}\label{fig4}       
\end{figure}
\begin{figure}
\resizebox{1.0\columnwidth}{!}{
  \includegraphics{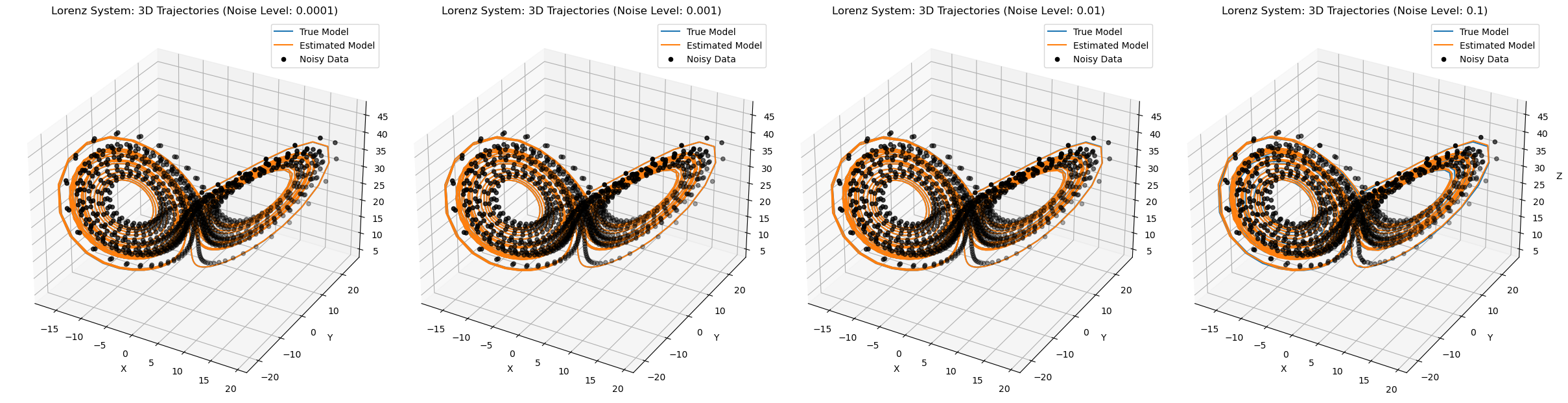} }
\caption{Comparing True and Identified Phase Portraits of the Lorenz System: This illustration offers a side-by-side examination of the true phase portrait of the Lorenz system, spanning the time span from $t=0$ to $t=25$ and initialized with $[x_{0} \ y_{0} \ z_{0}]^{T} = [-8 \ 7 \ 27]^{T}$, with the phase portrait of the identified systems under various degrees of Gaussian noise. The primary objective of this comparison is to evaluate the precision with which the identified systems replicate the intricate dynamics of the original system.}\label{fig5}       
\end{figure}
Figures \ref{fig4} and \ref{fig5} offer visual insights into the Lorenz system's dynamics and the accuracy of our estimation method. Figure \ref{fig4} highlights the dynamic trajectories of the Lorenz system, showcasing the influence of noise on position and velocity measurements. The solid blue lines represent true trajectories, while the dashed red arrows depict estimated trajectories obtained through neural networks. On the other hand, Figure \ref{fig5} provides a side-by-side comparison of true and identified phase portraits under varying levels of Gaussian noise, allowing us to assess the precision of our identification method.
\begin{table}[h!]
\caption{Impact of Noise Characteristics on Parameter Estimation and Solution Accuracy: A Comparative Analysis between Gaussian and Colored (Pink) Noise}\label{table8}
\centering
\begin{tabular}{ccc}
\hline
  & Gaussian Noise & Colored (Pink) Noise \\
\hline
$\hat{\sigma}$ & 9,9989 & 10,0015 \\
$\hat{\rho}$ & 27,9972 & 28,0042 \\
$\hat{\beta}$ & 2,6662 & 2,6668 \\
Huber Loss & 0,000126& 0,0002245\\
\hline
\end{tabular}
\end{table}
Table \ref{table8} facilitates a comparative analysis between Gaussian and colored (pink) noise, shedding light on the impact of noise characteristics on parameter estimation and solution accuracy. The estimated parameters remain consistent across both types of noise, with slightly improved accuracy observed for Gaussian noise.

\section{Conclusion}\label{sec13}
In this study, we thoroughly explored the use of a Huber loss-guided neural network,  employing a multilayer perceptron architecture, to effectively predict parameters across various nonlinear systems. Our primary focus centered on the accurate estimation of parameters within damped oscillators, van der Pol oscillators, Lotka-Volterra systems, and Lorenz systems under the influence of multiplicative noise. This noise encompassed both Gaussian and pink noise, each at varying intensities.
Our experiments have unveiled promising insights, effectively highlighting the robustness of our proposed approach in parameter estimation for these intricate nonlinear systems. The utilization of the multilayer perceptron, guided by the Huber loss, exhibited exceptional prowess in predicting parameters that capture the systems' dynamics. This highlights the ability of neural networks to uncover essential patterns within noisy data and extract valuable information about system parameters.
An important observation from our study was the inherent resilience of our approach across a diverse range of noise levels. Regardless of whether the noise took the form of Gaussian or pink noise, the neural network consistently produced parameter estimations that remained reliable, even in the presence of noise. This ability to adapt to various noise scenarios holds substantial relevance in practical scenarios where noise is an intrinsic aspect of real-world experimental data.

In summary, our study makes a valuable contribution to the evolving landscape of parameter estimation. We have effectively showcased the potential of Huber loss-guided neural networks, particularly those adopting the multilayer perceptron architecture, in accurately predicting parameters within complex nonlinear systems. The applicability of this approach extends seamlessly to damped oscillators, van der Pol oscillators, Lotka-Volterra systems, and Lorenz systems, all operating under the influence of multiplicative noise. This versatility underscores its relevance across various domain of scientific and engineering domains. Ultimately, our findings underscore the feasibility of harnessing machine learning techniques to elevate the accuracy of parameter estimation, even when confronted with intricate and noisy datasets.

\backmatter
\section*{Declarations}

\subsection*{Author Contribution:}
K.K. conceptualized the research, conducted experiments, analyzed data, and authored the manuscript.

\subsection*{Conflict of Interest:}
The author declares no competing interests.

\subsection*{ORCID iD:}
Kaushal Kumar: \url{https://orcid.org/0000-0002-2555-9623}

\subsection*{Funding:}
This study received no external funding.

\subsection*{Data Availability Statement/Code Availability:}
This study is centered around a theoretical exploration and, therefore, does not hinge on empirical or experimental data. The simulations in this paper utilize synthetic data created for this research. The code used for generating results will be made available on GitHub\footnote{\url{https://github.com/kaushalkumarsimmons/ML_parameter_estimation}} upon manuscript acceptance.





\bibliography{sn-bibliography}

\end{document}